\documentclass[12pt,onecolumn]{article}
\usepackage{amsmath,amsthm,amssymb}
\usepackage{euscript}
\usepackage{enumerate}
\usepackage[pdftex]{graphicx}

\usepackage{harvard}

\newcommand{\be}{\begin{equation}}
\newcommand{\ee}{\end{equation}}

 \title{A Semiparametric Bayesian Extreme Value Model Using a Dirichlet Process
 Mixture of Gamma Densities}

\author{Jairo A Fuquene P.
\thanks{
Department of Applied Mathematics and Statistics, Jack Baskin
School of Engineering University of California, Santa Cruz, USA.
jfuquene@soe.ucsc.edu}}

\begin{document}

\maketitle

\bibliographystyle{agsm}
\citationstyle{agsm}

\date

\begin{abstract}
In this paper we propose a model with a Dirichlet process mixture
of gamma densities in the bulk part below threshold and a
generalized Pareto density in the tail for extreme value
estimation. The proposed model is simple and flexible allowing us
posterior density estimation and posterior inference for high
quantiles. The model works well even for small sample sizes and in
the absence of prior information. We evaluate the performance of
the proposed model through a simulation study. Finally, the
proposed model is applied to a real environmental data.
\end{abstract}

\textbf{keywords} Generalized Pareto Distribution, Threshold
Estimation, Dirichlet Process Mixture.

\section{Introduction}

In recent years extreme value mixture models have been proposed as
a combination of a distribution with a ``bulk part" below
threshold and a generalized Pareto distribution (GPD) in the tail.
Different distributions have been proposed for modelling the
``bulk part" where the threshold is a parameter to be estimated.
The first approach which allow us a transition between the bulk
and tail parts is provided by \citeasnoun{friguessi}.
\citeasnoun{friguessi} uses a Weibull distribution in the bulk
part, a GPD for the tail and the location-scale Cauchy cdf in the
transition function and the authors use maximum likelihood
estimation. However in the \citeasnoun{friguessi} approach maximum
 likelihood estimation in the bulk
 part could produce multiple modes and hence some
 identifiability problems. \citeasnoun{Behrens} and \citeasnoun{carreu} consider Gamma
and Normal distributions respectively in the bulk part. But an
unimodal distribution is not realistic in practice where the
density has different unknown shapes in many applications.
\citeasnoun{gammerman} use Bayesian inference in the bulk part
following the proposal of \citeasnoun{wiper} who propose to assign
prior probabilities on the number of components of the mixture of
gammas and to use the reversible jump algorithm for posterior
inference purposes. The authors use BIC and DIC criteria for model
comparison on a fixed number of gamma components. This approach
allow us to have a flexible model with multimodality in the bulk
distribution. Also, \citeasnoun{gammerman} show that using
posterior predictive inference the discontinuity problem at the
threshold is eliminated. \citeasnoun{Macdonald} et al propose a
non-parametric approach in the bulk part with kernel bandwidth
estimators and a GPD in the tail where Bayesian inference is
applied. For a more exhaustive discussion of extreme value
threshold estimation see for example \citeasnoun{mamamia}. On the
other hand, there is an extensive literature on Dirichlet mixture
process for density estimation particulary using gaussian
distributions the main paper is given by \citeasnoun{escobar}. The
Dirichlet process is very flexible, theoretically coherent and
simple and in recent years it has been an important tool of many
applications for Bayesian density estimation
(\citeasnoun{Ferguson} and \citeasnoun{Antoniak}).
\citeasnoun{Hanson} proposes the Dirichlet process mixture of
gamma densities (DPMG) for density estimation of univariate
densities on the positive real line.\\

In this paper we propose a model with a  DPMG below threshold and
a GPD in the tail. We have important reasons for using the
proposed model: First, because DPMG could be a powerful tool for
density estimation in the bulk part  (allow us accommodate a very
wide variety of shapes and spreads in the bulk part) the tail fit
is expected to be adequate. Second, the proposed model can be used
in the absence of prior information. Third, Dirichlet Process
Mixture controls the expected number of components
(\citeasnoun{Antoniak}) therefore the extensive task for model
comparison purposes using BIC and DIC criteria on a fixed number
of gamma components in the bulk part is not necessary. In
addition, because DPMG is random we can build credible intervals
of the posterior density in the bulk part. This paper is organized
as follows. Section 2 is devoted to present the proposed model. In
Section 3 we present a simulation study of the proposed model. In
Section 4 the proposed model is applied to a real environmental
data. Finally in Section 5 we have the conclusions.

\section{Model}

The density of the Generalized Pareto Distribution with scale
parameter $\sigma$ and shape parameter $\gamma$ is as follows:
\begin{equation}
g(x|\phi)=
  \begin{cases}
   \frac{1}{\sigma}\left(1+\xi\frac{(x-u)}{\sigma}\right)^{-(1+\xi)/\xi} & \text{if}  \;\;\; \xi\neq 0\\
   \frac{1}{\sigma}\exp(-(x-u)/\sigma) & \text{if} \;\;\; \xi= 0,\\
  \end{cases}
\end{equation}

where the vector of parameters $\phi=(\xi,\sigma,u)$ and $x-u>0$
for $\xi\geq 0$ and $0\leq x-u<-\phi/\xi$ for $\xi< 0$. We have
that GDP is bounded from below by $u$, bounded from above by
$u-\sigma/\xi$ if $\xi<0$ and unbounded from above if $\xi\geq 0$.
The density of the proposed model is the following:

\begin{equation}\small
f(x|\phi,\theta)=
  \begin{cases}
   h(x|\theta) & \;\; x\leq u\\
   [1-H(u|\theta)]g(x|\Phi) & \;\; x> u\\
  \end{cases}
\end{equation}

where $\phi=(u,\xi,\lambda)$, $\theta=(\lambda,\gamma)$ and
$H(u|\theta)$ denotes the cumulative distribution function (CDF)
of $h(x|\theta)$ at $u$. The cumulative distribution function of
(2) is as follows:
\begin{equation}\small
F(x|\phi,\theta)=
  \begin{cases}
   H(x|\theta) & \;\; x\leq u\\
   H(u|\theta) + [1-H(u|\theta)]G(x|\phi) & \;\; x> u\\
  \end{cases}
\end{equation}
where $G(x|\phi)$ is the CDF of GPD. Note that $_{\lim
x\longrightarrow u^{-}}F(x|\phi,\theta) =H(u|\theta)$ and
\linebreak $_{\lim x\longrightarrow u^{+}}F(x|\phi,\theta)
=H(u|\theta)$ therefore (3) is continuous at $u$.

\subsection{The Dirichlet Process Mixture of Gamma densities}

The novel proposal is to use in the bulk part of (2) a DPMG, as
follows we have a short introduction about the DP. A distribution
G on $\Theta$ follows a dirichlet process $DP(\alpha,G_{0})$ if,
given an arbitrary measurable partition, $B_{1},B_{2},...,B_{k}$
of $\Theta$ the joint distribution of
$(G(B_{1}),G(B_{2}),...,G(B_{k}))$ is Dirichlet $(\alpha
G_{0}(B_{1}),\alpha G_{0}(B_{2}),...,\alpha G_{0}(B_{k}))$ where
$G(B_{i})$ and $G_{0}(B_{i})$ denote the probability of set
$(B_{i})$ under $G$ and $G_{0}$ respectively (see
\citeasnoun{Ferguson}). Here $G_{0}$ is a specific distribution on
$\Theta$ and $\alpha$ is a precision parameter. Let
$K(;,\boldsymbol{\theta})$ be a parameter family of distributions
functions (CDF's) indexed by $\boldsymbol{\theta}\; \;\epsilon\;\;
\Theta$, with associated densities $k(;\boldsymbol{\theta})$. If
$G$ is proper we define the mixture distribution
\begin{equation}
F(.;G)=\int K(;,\theta)G(d\boldsymbol{\theta})
\end{equation}
where $G(d\boldsymbol{\theta})$ can be interpreted as the
conditional distribution of ${\theta}$ given $G$. We can express
(4) as $f(.;G)=\int k(.;G)$ differentiating with respect to $(.)$.
Due to $G$ is random then $F(.;G)$ is random. $F(.;G)$ is the
model for the stochastic mechanism corresponding to
$x_{1},x_{2},...,x_{n}$ assuming $x_{i}$ given $G$ are i.i.d. from
$F(.;G)$ with the DP structure. In this paper we implement the
Dirichlet Process Mixture model by using the P\'olya urn scheme
 (see \citeasnoun{escobar}
and \citeasnoun{miki}). In DPMG we have mixing parameters
$(\lambda_{i},\gamma_{i})=$ associated with each $x_{i}$. The
model can be expressed in hierarchical form as follows:
\begin{align}
x_{i}&|\lambda_{i},\gamma_{i} \sim h(x_{i},\theta_{i}), \;\;\; i=1,..,n\\
\notag \theta_{i}|G &\sim G, \;\;\; i=1,..,n
\\ \notag G&|\alpha,\eta
 \sim DP(\alpha,G_{0}), G_{0}=G_{0}(.|\eta) \\
\notag \alpha,\eta
 &\sim p(\alpha)p(\eta)
\end{align}
here $\theta_{i}=(\lambda_{i},\gamma_{i})$ and
$h(x_{i},\theta_{i})$ denotes the gamma density with the scale
parameter, $\lambda_{i}$ , and the shape parameter, $\gamma_{i}$,
\begin{equation}
h(x_{i}|\lambda_{i},\gamma_{i})=\frac{\gamma_{i}^{\lambda_{i}}}{\Gamma(\gamma_{i})}x_{i}^{\lambda_{i}-1}\exp\left\{-\gamma_{i}
x_{i}\right\} \;\; x_{i}>0
\end{equation}
We use the approach of \citeasnoun{Hanson} for
$g_{0}(\lambda,\gamma|\eta)$ therefore two independent exponential
distributions are considered as follows
\begin{equation}
g_{0}(\lambda,\gamma|\eta)=a_{\lambda}\exp(-a_{\lambda}\lambda)a_{\gamma}\exp(-a_{\gamma}\gamma)
\end{equation}
with $\eta=(a_{\lambda},a_{\gamma})$. The parameters of (7) follow
gamma priors $a_{\lambda}\sim \Gamma(b_{\lambda}, c_{\lambda})$
and $a_{\gamma}\sim \Gamma(b_{\gamma}, c_{\gamma})$, where
$\Gamma(a, b)$ denotes the gamma density with parameters $a$ and
$b$.

\subsection{Priors for the parameters in the generalized Pareto distribution}

Now we present the priors for the threshold $u$, the scale
parameter $\sigma$ and shape parameter $\xi$ of the GPD. The prior
distribution for $u$ is a normal density $N(m_{u},\sigma^{2}_{u})$
as suggested \citeasnoun{Behrens}. \citeasnoun{catellanos} obtain
the Jeffrey's non-informative prior for $(\sigma,\xi)$ and the
authors show this prior produces proper posterior results. The
prior is the following:

\begin{equation}
p(\sigma,\xi)\propto \sigma^{-1}(1+\xi)^{-1}(1+2\xi)^{-1/2}
\end{equation}
 where $\xi>-0.5$ and $\sigma>0$. According to
\citeasnoun{coles} situations were $\xi<-0.5$ are very unusual in
practice. The posterior distribution on the log-scale using the
density (2) is then:

\clearpage

\begin{figure}

\begin{align}
\log(p(\theta,\Phi|x))&\propto
\sum_{A}\log(h(x|\theta))+\sum_{B}\log\left((1-H(u|\theta))
\frac{1}{\sigma}\left(1+\xi\frac{(x-u)}{\sigma}\right)^{-(1+\xi)/\xi}\right)\\
&+ \log(p(u)p(\xi)p(\sigma)) \notag
\end{align}
for $\xi\neq 0$ and
\begin{align}
\log(p(\theta,\Phi|x))&\propto
\sum_{A}\log(h(x|\theta))+\sum_{B}\log\left((1-H(u|\theta))
\frac{1}{\sigma}\exp(-(x-u)/\sigma)\right)\\
&+ \log(p(u)p(\xi)p(\sigma)) \notag
\end{align}
for $\xi= 0$. With $A=\left\{x_{i}:x_{i}\leq u\right\}$ and
$B=\left\{x_{i}:x_{i}>u\right\}$.
Using the proposed model we can compute high quantiles below
threshold. In order to find values beyond the threshold we have
that
\begin{equation}
F(x|\phi,\lambda,\gamma)=H(u|\lambda,\gamma)+[1-H(u|\lambda,\gamma)]
G(x|\phi)
\end{equation}
where $G(x|\phi)$ is the CDF of the GPD. For example to find the
$p$-quantile, $q$, we use
\begin{equation}
p^{*}=\dfrac{p-H(u|\lambda,\gamma)}{1-H(u|\lambda,\gamma)}
\end{equation}
and solve $G(q|\phi)=p^{*}$.
\begin{center}
\includegraphics[scale=0.4]{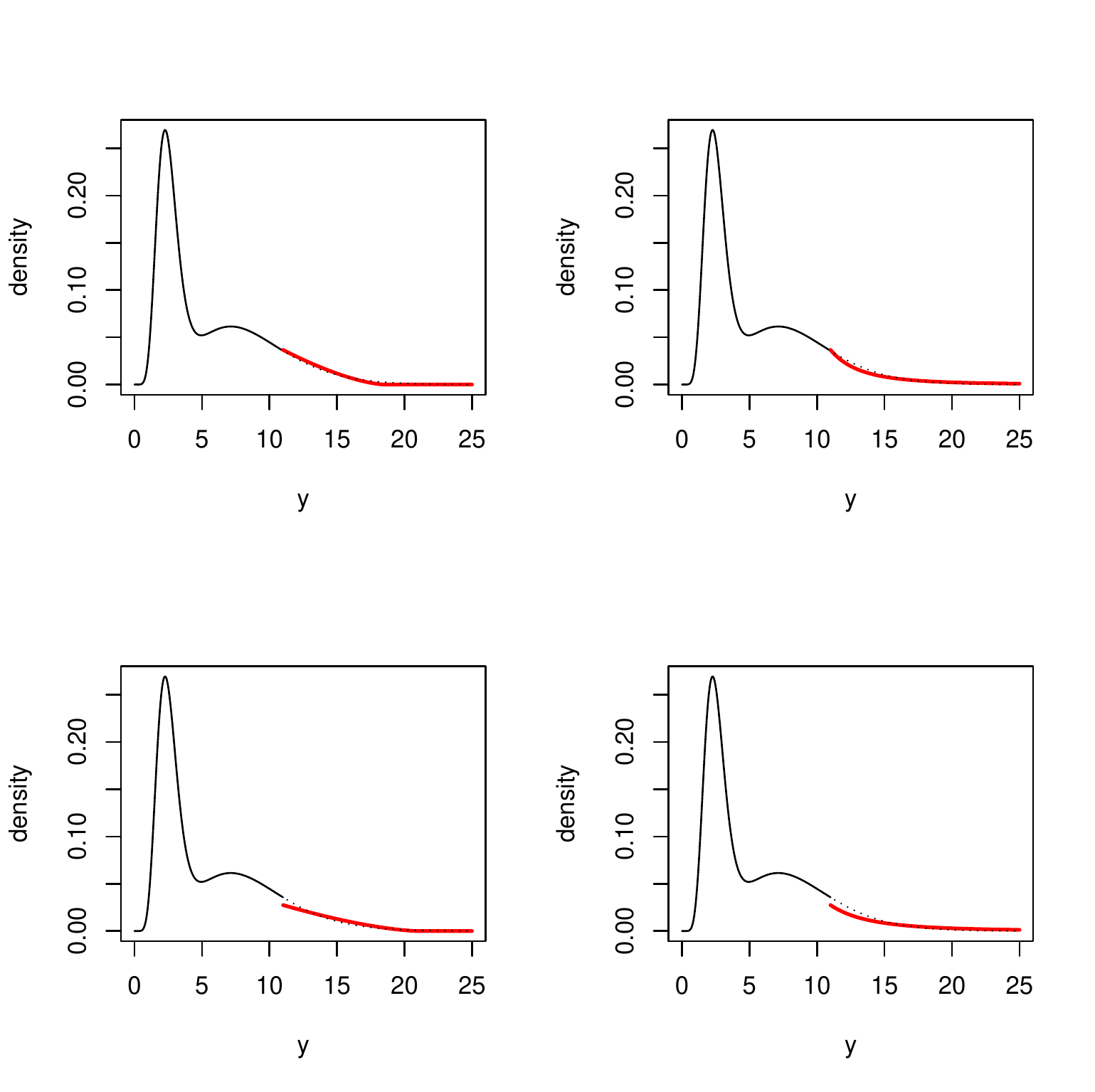}
\caption{Probability density function of the model (3) for a
number of parameters values: (a) $\xi=-0.4$ and $\sigma=3$, (b)
$\xi=0.4$ and $\sigma=3$, (c) $\xi=-0.4$ and $\sigma=4$ and  (d)
$\xi=0.4$ and $\sigma=4$, threshold $u=11$ and the center of the
densities is a mixture of two gamma densities the tails are
modelling with GPD.}
\end{center}
\end{figure}

\clearpage

Figure 1. displays the density of the proposed model considering
different values in the parameters. This model allows a
discontinuity of the density at the threshold because constrains
are basically solve defining the adequate models we consider in
this paper for the posterior analysis  in the tail (see
\citeasnoun{gammerman}).

\section{Simulation study}

In this section we evaluate the performance of the proposed model
through a simulation study. The precision $\alpha$ of $g_{0}$ in
the DP affects the expected number of components in the mixture.
\citeasnoun{Hanson} consider values of $\alpha$ fixed to 0.1 and 1
and also random values using different assignments of Gamma priors
for $\alpha$ such as $\Gamma(2,2)$ and $\Gamma(2,0.5)$. Here we
consider the precision for DP using $\alpha=0.1$. The
hyperparameters of $g_{0}$ can be expressed in terms of the mean
$\mu=\lambda/\gamma$ and variance $V=\lambda/\gamma^{2}$ of
$h(x|\theta)$ (see \citeasnoun{Hanson}) as two diffuse densities
$f(\mu|a_{\lambda},a_{\gamma})=a_{\lambda}a_{\gamma}/(a_{\lambda}\mu+a_{\gamma})^{2}$
and
$f(V^{-1}|a_{\lambda},a_{\gamma})=\Gamma(2,a_{\lambda}\mu^{2}+a_{\gamma}\mu)$
respectively. Suppose now that $a_{\lambda}=a_{\gamma}=1$, so
$f(\mu|1,1)=1/(1+\mu)^{2}$ which is the Beta Prime distribution
with hyperparamters of scale and shape equals to 1. The Beta prime
distribution has been used as default density for modelling
inference in Bayesian analysis. Therefore we can think that we are
modelling the mean of the mixture of gammas densities in a non
informative (but robust) manner.  We consider a small sample size
$n=200$. \citeasnoun{Hanson} obtain an accurate smooth in an
univariate density using DPMG with different specifications for
$\alpha$ and large sample sizes 1000 and 10000. Here, we have that
$\xi=0.4$, $\sigma=3$ and the threshold $u=11$ at the 90\%
quantile in the simulated data. The simulated mixture density for
the central part is:
\begin{equation}
h(x)=0.5\Gamma(x|10,4)+0.5\Gamma(x|6,0.7).
\end{equation}
Following \citeasnoun{Hanson}  the hyperparameters for
$a_{\lambda}$ and $a_{\gamma}$ are
$b_{\lambda}=b_{\gamma}=c_{\lambda}=c_{\gamma}=0.001$ in order to
have a non informative $g_{0}$. The prior of the threshold $u$ has
mean equal to 90\% quantile in the simulated data  and the
variance $\sigma^{2}_{u}$ gives 99\% of probability in the range
between 50\% and 99\% of the simulated data. As usual in the
Metropolis algorithm, we adjust the variance of the sampling
proposal densities considering the hessian of the maximum
likelihood estimates using some MCMC simulations. We obtained
convergence of all parameters using 10000 iterations after a
burn-in period of 5000 iterations.  Figures 2 displays the quality
of the approach even with a small sample size of $n=200$. The
posterior density in the proposed model reproduces the underline
density with precision according to the credible interval in the
bulk part and posterior predictive mean in the tail. The density
estimation in bulk part of the proposed model could be even better
when large sample sizes are considered (see \citeasnoun{Hanson}).
Figure 3 displays the posterior densities of threshold $u$, scale
$\sigma$, and shape $\xi$. We can see the posterior distribution
represents nicely the true parameters. In particular the threshold
is centered around the true value 11. Figures 4 shows as the
posterior distributions of the predictive quantiles at 95\% is
accurately estimated.

\clearpage

\begin{figure}[ht]
\begin{center}
\includegraphics[scale=0.9]{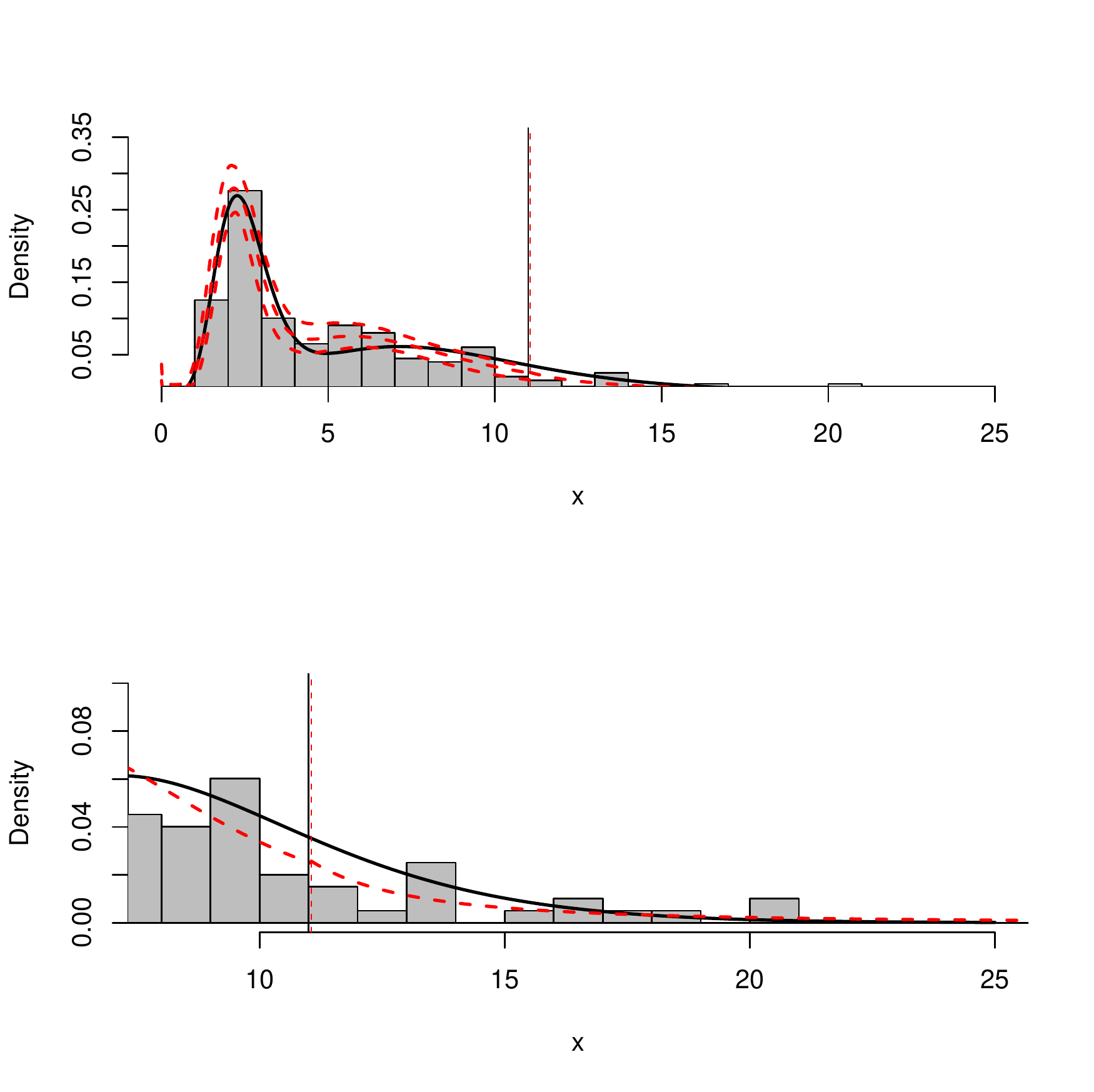}
\caption{Dashed red line: Posterior predictive density using the
Dirichlet process mixture of gamma densities in the bulk part and
a GPD in the tail. Full black line is the true density and the
dashed red line is the simulated density. The histogram displays
the simulated data. The vertical full black line is the true
threshold location and the vertical dashed red line is the
posterior threshold location.}
\end{center}

\end{figure}

\clearpage

\begin{figure}[ht]

\begin{center}
\includegraphics[scale=0.5]{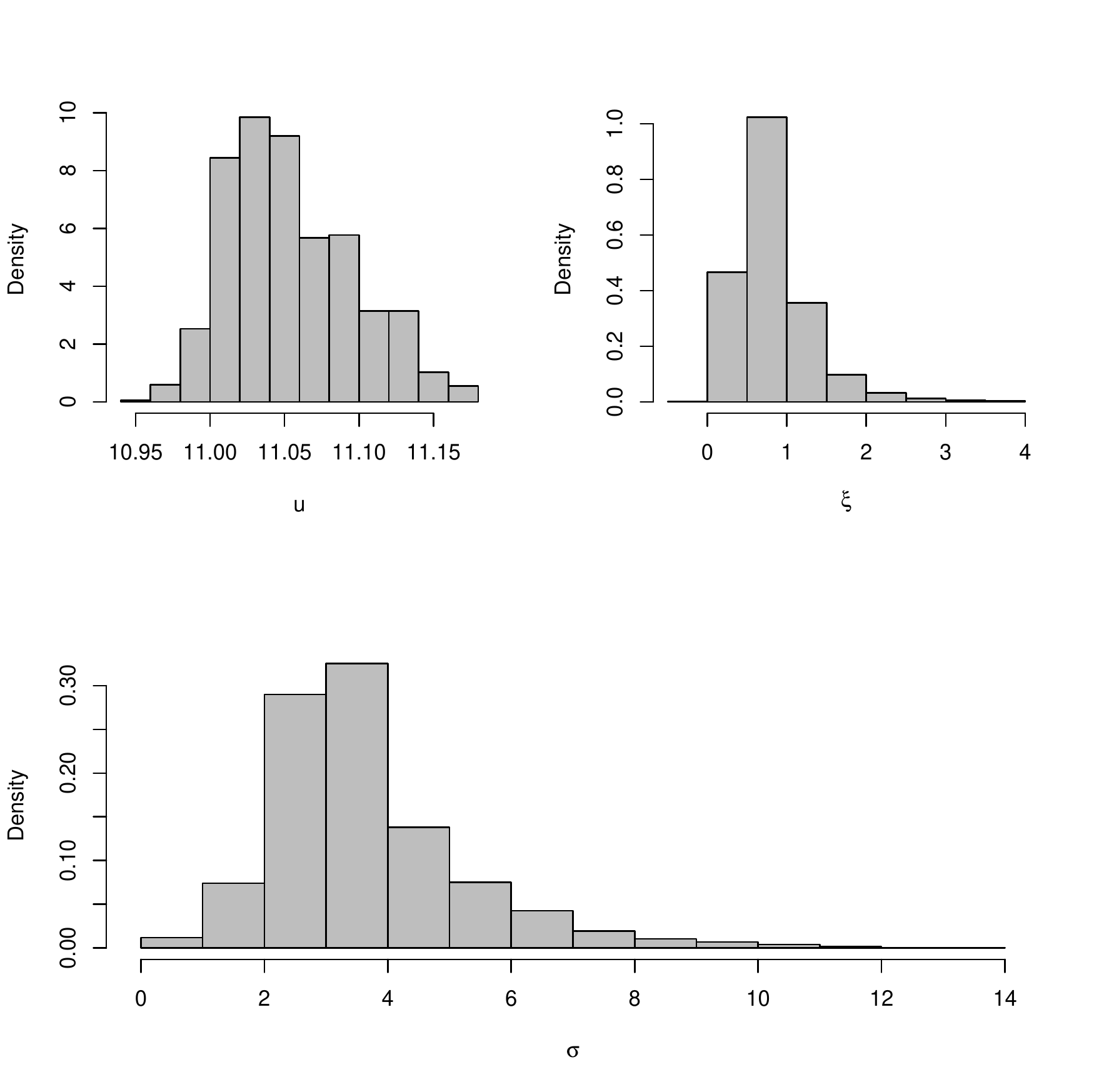}
\caption{Posterior distribution of $u$, $\xi$ and $\sigma$.}
\end{center}

\begin{center}
\includegraphics[scale=0.5]{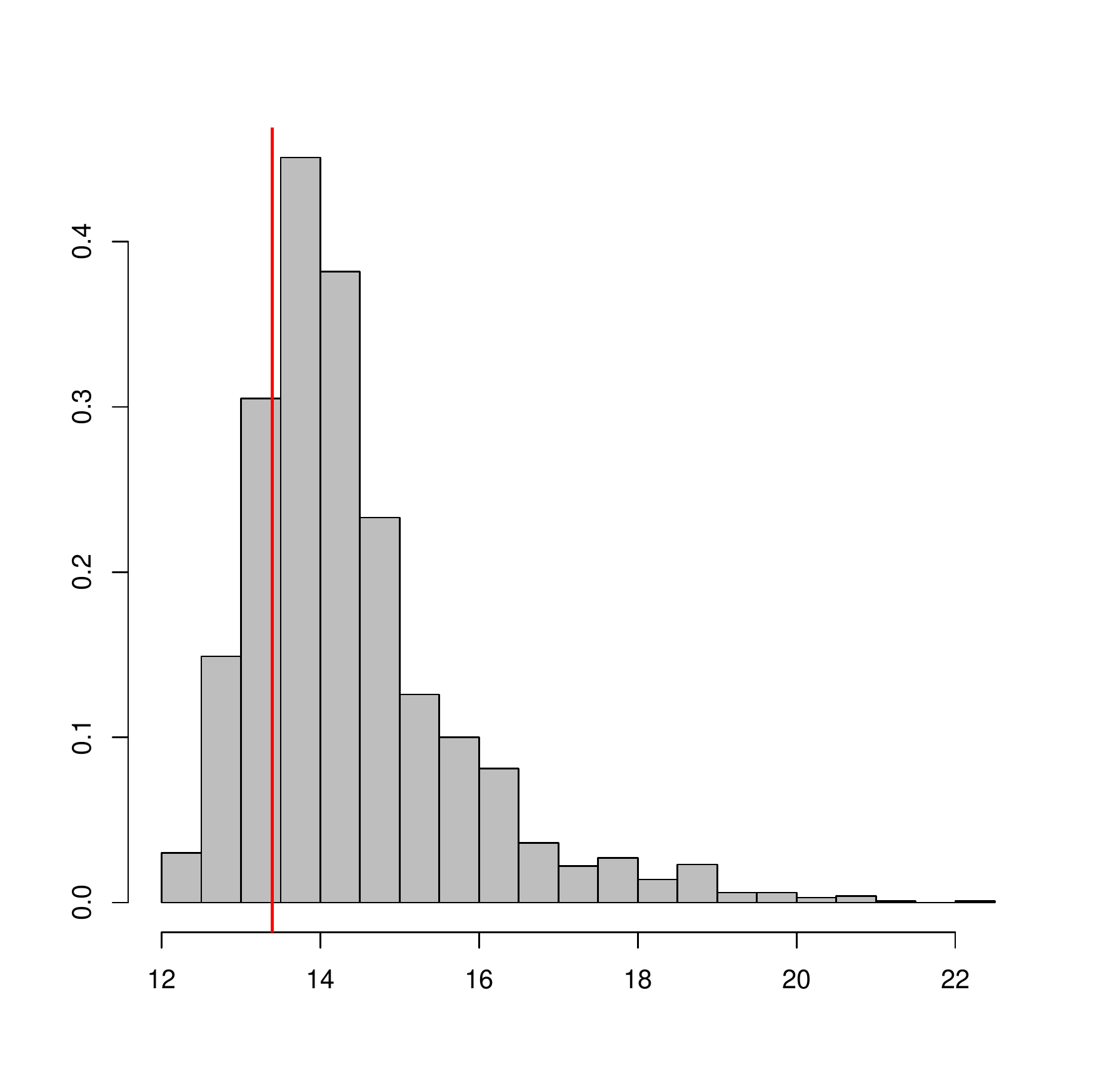}
\caption{Posterior histogram of the 95\% quantile for the
simulation. Red line the true quantile.}
\end{center}

\end{figure}

\clearpage

\begin{figure}[ht]

\section{Application to the flow levels in the Gurabo river}

River flow levels are important measures to prevent disasters in
populations when flow rate exceeds the capacity of the river
channel. We applied the proposed model in river flow levels
measured at cubic feet per second ($ft^{3}/s$) in Gurabo river at
Gurabo Puerto Rico. The data is available at waterdata.usgs.gov.
The flows are monitoring between December 2 2012, 12:00 am to
December 4 2012, 8:45 pm. The measures are made each 15 minutes
for a total sample size of n=254. We obtained convergence of all
parameters using 5000 iterations after a burn-in period of 2000
iterations.

Figure 5 displays the posterior distributions of the parameters in
the tail of the proposed model. The threshold, scale and shape are
around of 1430 (quantile at 96\% according to the simulation), 300
and -0.25. Figure 6 shows the posterior distribution for the
99.9\% high quantile, we can see the maximum value is less than
the posterior mean for the quantile at 99.9\% and the posterior
distribution is asymmetric which is expected. Figure 7 displays
the posterior density using DPMG in the bulk part and a GPD in the
tail. We can see our proposed model reproduces the data in the
bulk and tail parts. As a conclusion according to the posterior
analysis (based on the last two days) with a probability of 0.1\%
we can see values bigger than approximately 1998 $ft^{3}/s$ in the
Gurabo River.

\begin{center}
\includegraphics[scale=0.5]{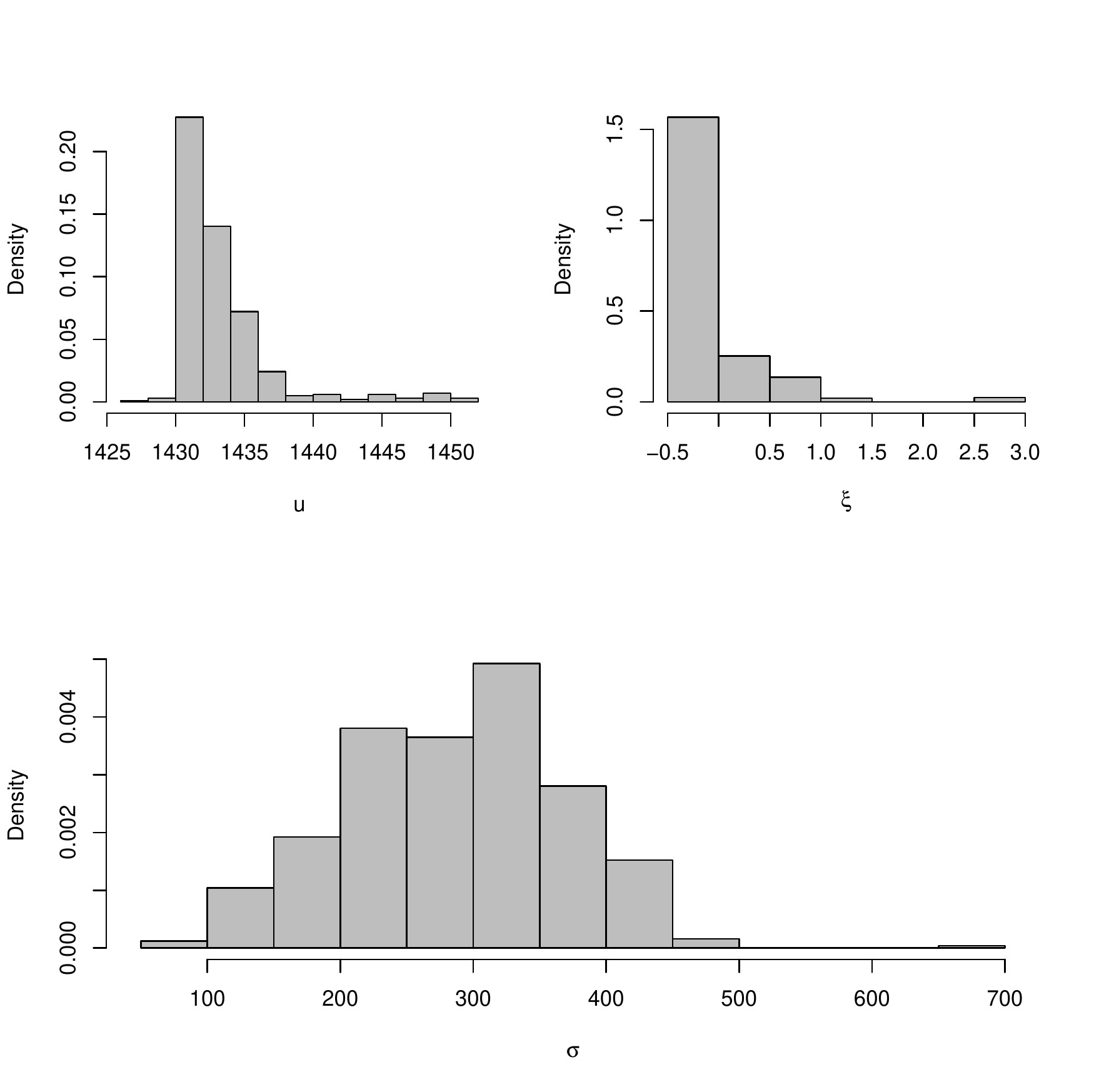}
\caption{Posterior histogram of the GPD parameters in the tail of
the proposed model for the application.}
\end{center}

\end{figure}

\clearpage

\begin{figure}[ht]
\begin{center}
\includegraphics[scale=0.5]{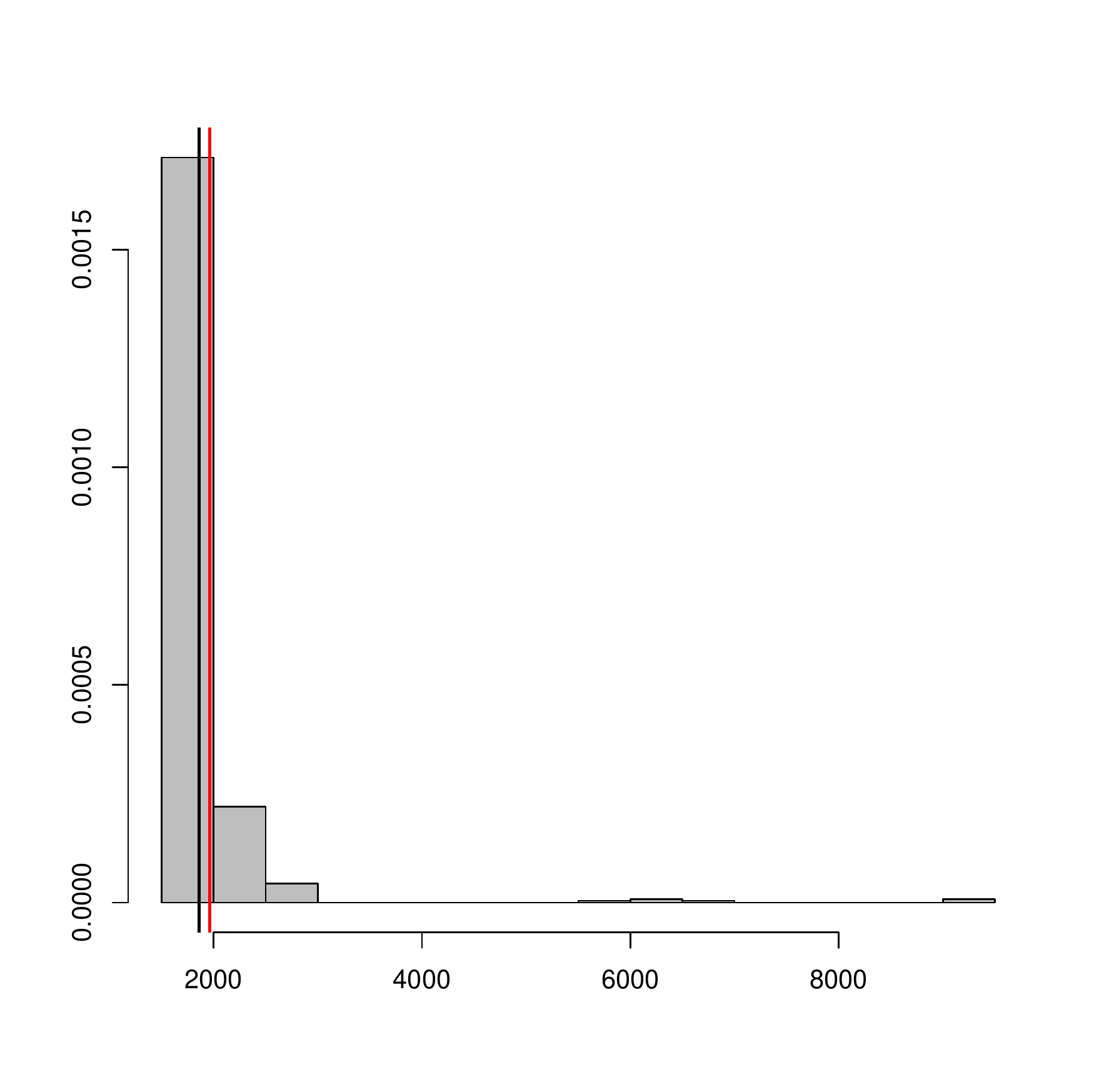}
\caption{Posterior distribution of the 99.9\% quantile for the
application. Black line the maximum observed data and red line the
posterior mean for the 99.9\% simulated quantile.}
\end{center}

\begin{center}
\includegraphics[scale=0.5]{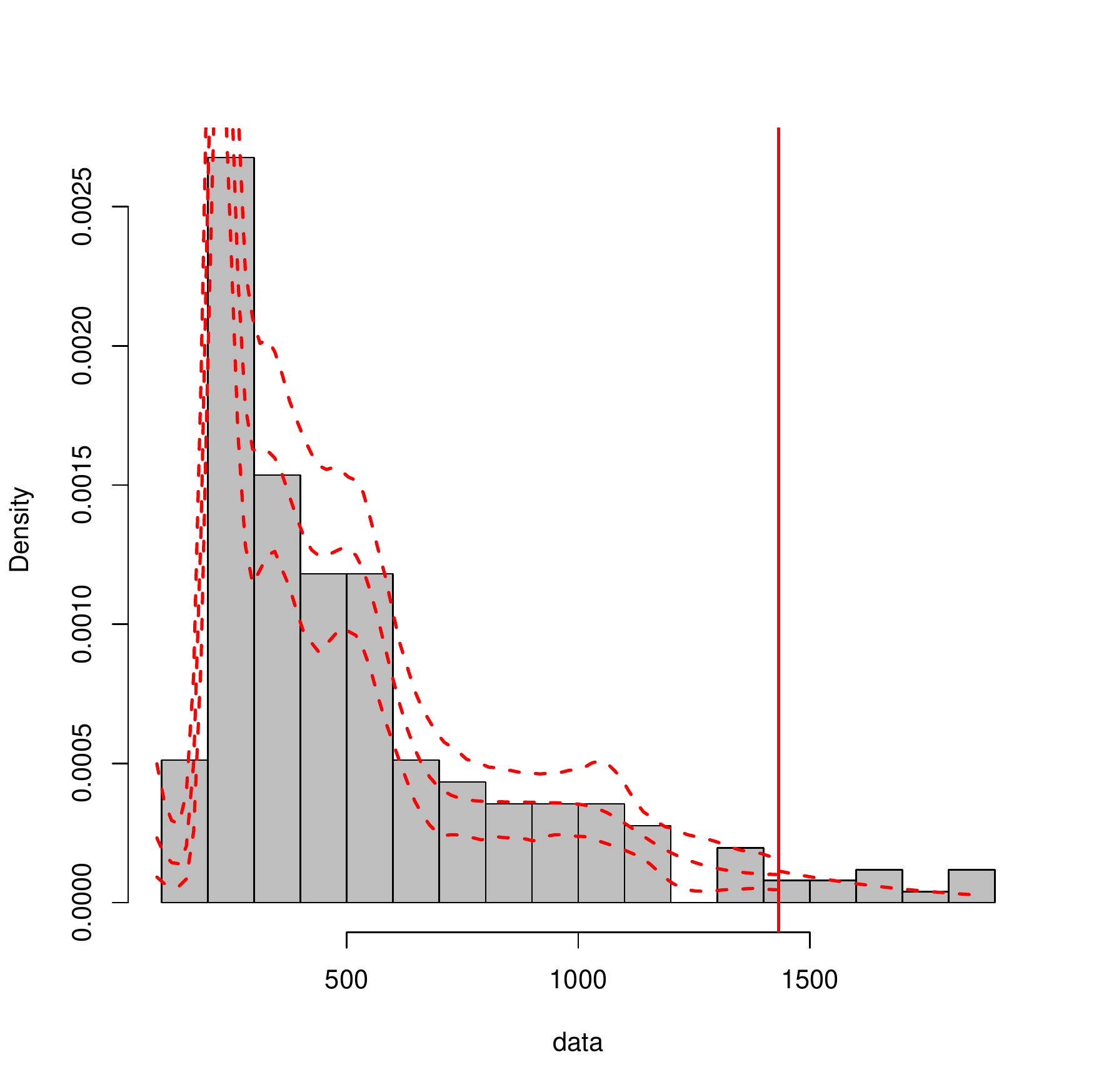}
\caption{Posterior density using the Dirichlet process mixture of
gamma densities in the bulk part and posterior predictive
distribution using a GPD in the tail. The vertical red line is the
posterior threshold location.}
\end{center}

\end{figure}

\clearpage

\section{Conclusion}

We proposed a model with a Dirichlet process mixture of gamma
densities in the bulk part of the distribution and a heavy tailed
generalized Pareto distribution in the tail for extreme value
estimation. The proposal is very flexible and simple for density
estimation in the bulk part and posterior inference in the tail.
According to the simulations and application to real data the
model works well even for small sample sizes and in the absence of
prior information. The Dirichlet Process mixture controls the
expected number of components and so the extensive task for model
comparison purposes using BIC and DIC criteria on a fixed number
of gamma components in the bulk part is not necessary. The
proposed model was applied to a real environmental data set but
interesting applications can be found in different areas such as
clinical trials or finance.

\bibliography{biblio}

\appendix

\section{MCMC algorithm}

\begin{enumerate}

\item For the bulk part we need to compute
$h(x|\theta)$ and also $H(u|\theta)$, we consider the p\'olya urn
expression in the DPMG to compute posterior realizations for the
density $h(x|\theta)$. Let
$\{\theta_{1}^{*},...,\theta^{*}_{n^{*}}\}$ the unique values of
$\theta_{i}$, $\omega_{i}=j$ if and only if
$\theta_{i}=\theta_{j}^{*}$ $i=1,2,,...,n$ and $n _{j}=|\{i:
\omega_{i}=j\}|$ and $j=1,2,...,n^{*}$ with $n^{*}$ number of
distinct values. We use the following transition probabilities:

\begin{enumerate}
  \item P\'olya urn: marginalized $G$ (using $^{-}$ to indicate summaries without
$\omega_{i}$) and defining a specific configuration
$\{\omega_{1},..,\omega_{n}\}$ with transition probabilities:

\begin{equation}
p(\omega_{i}=\ell|\omega_{-i})\propto
\begin{cases}
n_{j}^{-} & j=1,...,n^{*-},\\
\alpha  & j=n^{*-}+1
\end{cases}
\end{equation}
  \item Resampling cluster membership indicators $\omega_{i}$:

\begin{equation}\small
p(\omega_{i}=j|,...,x_{i})\propto
  \begin{cases}
n_{j}^{-}k(x_{i};\theta_{j}^{*}) & j=1,...,n^{*-}, \\
\alpha\int k(x_{i};\theta_{i})dG_{0}(\theta_{i}|\eta) & j=n^{*-}+1
  \end{cases}
\end{equation}

where we use the close results in \citeasnoun{Hanson}:

\begin{align}
k(x_{i};\theta_{j}^{*})&=h(x_{i}|\theta_{j}^{*})
\end{align}

\begin{align} \small
\int k(x_{i};\theta_{i})dG_{0}&(\theta_{i}|\mu,\tau^{2})=\\
&\frac{a_{\lambda}a_{\gamma}}{x_{i}(x_{i}+a_{\lambda})(a_{\lambda}-\log(x_{i}/(x_{i}+a_{\gamma})))^{2}}
\end{align}

with probability proportional to
$n_{j}^{-}k(x_{i};\theta_{j}^{*})$ we make
$\theta_{i}=\theta_{j}^{*-}$. On the other hand with probability
proportional to $\alpha\int
k(y_{i};\theta_{i},\phi)dG_{0}(\theta_{i}|\eta)$ we open a new
component and we sample $\theta_{i}=(\lambda_{i},\gamma_{i})$.
First we sample $\lambda_{i}|\eta \sim
\Gamma(2,a_{\lambda}-\log(x_{i}/(x_{i}+a_{\gamma}))^{2})$ then we
sample $\gamma_{i}|\lambda_{i},\eta \sim
\Gamma(\lambda_{i}+1,x_{i}+a_{\gamma})$.

\end{enumerate}

\item Now we are interested in to show the sampling for the
parameters in the GPD defined in the tails of (2). Following
\citeasnoun{gammerman} we compute the posterior distribution of
$u$, $\xi$ and $\sigma$ using three steeps of the Metropolis
Hasting algorithm. The algorithm is as follow:

\begin{enumerate}
\item Sampling $\xi$: proposal transition kernel is given by a
  truncated normal

\begin{equation}
\xi^{*}|\xi^{b} \sim
N(\xi^{s},V_{\xi})I(-\sigma^{b}/(M-u^{b}),\infty)
\end{equation}

where $V_{\xi}$ is a variance in order to improve the mixing. $M$
is the maximum value in the sample the acceptance probability is

\begin{equation}\small \notag
\alpha_{\xi}=\min\left\{1,\dfrac{p(\theta^{*},\phi^{*}|x)\Phi((\xi^{b}+\sigma^{b}/(M-u^{b}))/\sqrt{V_{\xi}})}{
p(\theta^{b},\phi^{b}|x)\Phi((\xi^{*}+\sigma^{*}/(M-u^{*}))/\sqrt{V_{\xi}})}\right\}
\end{equation}

where is the density function of the standard normal distribution.

  \item Sampling $\sigma$: If $\xi^{(b+1)}>0$ then $\sigma^{*}$
  is sampled from the Gamma distribution $\Gamma(\sigma^{2(b)}/V_{\sigma},\sigma^{b}/V_{\sigma})$
     where $V_{\sigma}$ is a variance in order to improve the mixing.
     On the other hand if $\xi^{(b+1)}<0$ then $\sigma^{*}$ is sampled from
     a truncated normal
\begin{equation}
\sigma^{*}|\sigma^{b} \sim
N(\sigma^{s},V_{\sigma})I(-\xi^{(b+1)}(M-u^{b}),\infty)
\end{equation}

the acceptance probabilities are respectively:

\begin{equation}\small \notag
\alpha_{\sigma}=\min\left\{1,\dfrac{p(\theta^{*},\phi^{*}|x)\Phi((\sigma^{b}+\xi^{(b+1)}(M-u^{b})/\sqrt{V_{\sigma}})}{
p(\theta^{b},\phi^{b}|x)\Phi((\sigma^{*}+\xi^{(b+1)}(M-u^{b})/\sqrt{V_{\sigma}})}\right\}
\end{equation}

and

\begin{equation}\small \notag
\alpha_{\sigma}=\min\left\{1,\dfrac{p(\theta^{*},\phi^{*}|x)\Gamma(\sigma^{b}|\sigma^{2(*)}/V_{\sigma},\sigma^{*}/V_{\sigma})}{
p(\theta^{b},\phi^{b}|x)\Gamma(\sigma^{*}|\sigma^{2(b)}/V_{\sigma},\sigma^{(b)}/V_{\sigma})}\right\}
\end{equation}

  \item The threshold $u^{*}$ is sampled following the
  requirement of the lower truncation for the GPD. Therefore
  $u^{*}$ is sampled using a truncated normal density

\begin{equation}
\sigma^{*}| \sigma^{b} \sim N(u^{s},V_{u})I(a^{(b+1)},\infty)
\end{equation}

If $\xi^{(b+1)}\geq 0$ then $a^{(b+1)}$ is the minimum value at
the sample in the iteration $b+1$ otherwise if $\xi^{(b+1)}<0$
$a^{(b+1)}=M+\sigma^{(b+1)}/\xi^{(b+1)}$. The acceptance
probability is then

\begin{equation}\small \notag
\alpha_{\xi}=\min\left\{1,\dfrac{p(\theta^{*},\phi^{*}|x)\Phi((u^{b}-a^{b+1})/\sqrt{V_{u}})}{
p(\theta^{b},\phi^{b}|x)\Phi((u^{b}-a^{b+1})/\sqrt{V_{u}})}\right\}
\end{equation}

\end{enumerate}

\end{enumerate}

\end{document}